\RequirePackage{fix-cm}
\documentclass[smallextended]{svjour3}       
\smartqed  
\usepackage{graphicx}
\usepackage{lipsum}
\usepackage{amsmath}
\usepackage{amsfonts}
\usepackage{booktabs} 
\usepackage{multirow}
\usepackage{graphicx}
\usepackage{xcolor}
\usepackage{epstopdf}
\usepackage{url}
\makeatletter
\def\cl@chapter{\@elt {theorem}}
\makeatother

\usepackage{todonotes}

\usepackage[capitalise]{cleveref}
\usepackage{algorithmic}
\usepackage{subcaption}

\usepackage{tikz}
\usetikzlibrary{external}
\tikzexternaldisable

\usepackage{pgfplots}
\pgfplotsset{compat=1.6}

\usepackage[backend=biber]{biblatex}
\ifpdf
\DeclareGraphicsExtensions{.eps,.pdf,.png,.jpg}
\else
\DeclareGraphicsExtensions{.eps}
\fi
\usepackage{array} 

\newcommand\numberthis{\addtocounter{equation}{1}\tag{\theequation}}

\newcommand{\R}{\ensuremath{\mathbb{R}}}

\def\abs#1{\left|#1\right|} 
\def\norm#1{\left\|#1\right\|} 

\newcommand{\Ls}{\ensuremath{{L}}_{\textrm{sym}}}

\newcommand{\eps}{\varepsilon}

\DeclareMathOperator*{\mingamma}{min_{\gamma}}

\newcommand{\bx}{{\bf x}}
\newcommand{\by}{{\bf y}}
\newcommand{\bxt}{\mathbf{\tilde{x}}}

\DeclareMathOperator{\dist}{dist}

\DeclareMathOperator{\DTW}{\mathrm{DTW}}
\DeclareMathOperator{\SDTWgamma}{\mathrm{DTW}_{\gamma}}
\DeclareMathOperator{\MPdist}{MPdist}

\definecolor{plotColorNearestNeighbor}{rgb}{0.0, 0.5, 0.0}
\definecolor{plotColorLinearSystem}{rgb}{0.8, 0.6, 0.0}
\definecolor{plotColorGCN}{rgb}{1.0, 0.0, 0.3}
\definecolor{plotColorAllenCahn}{rgb}{0.1, 0.6, 0.9}

\definecolor{darkgreen}{rgb}{0.0, 0.5, 0.1}
\definecolor{amber}{rgb}{0.9, 0.5, 0.0}

\newcommand{\change}[1]{\textcolor{black}{#1}}

\begin{document}
	
	\title{An Empirical Study of Graph-Based Approaches for Semi-Supervised Time Series Classification}

	
	\titlerunning{Graph-Based Semi-Supervised Time Series Classification}        
	
	\author{Dominik Alfke         \and
		Miriam Gondos \and
		Lucile Peroche \and 
		Martin Stoll
	}
	
	
	\institute{D. Alfke, M. Gondos, L. Peroche, M Stoll \at
		Chair of Scientific Computing, Department of Mathematics\\
		TU Chemnitz, Reichenhainer Str. 41, 09126 Chemnitz, Germany\\
		\email{\{dominik.alfke, lucile.peroche, martin.stoll\}@mathematik.tu-chemnitz.de}           
	}
	
	\date{Received: date / Accepted: date}

	\maketitle
	
	\begin{abstract}
		Time series data play an important role in many applications and their analysis reveals crucial information for understanding the underlying processes. 
		Among the many time series learning tasks of great importance, we here focus on semi-supervised learning based on a graph representation of the data.
		Two main aspects are involved in this task. A suitable distance measure to evaluate the similarities between time series, and a learning method to make predictions based on these distances.
		However, the relationship between the two aspects has never been studied systematically in the context of graph-based learning.
		We describe four different distance measures, including (Soft) DTW and MPDist, a distance measure based on the Matrix Profile, as well as four successful semi-supervised learning methods, including the graph Allen--Cahn method and a Graph Convolutional Neural Network.
		We then compare the performance of the algorithms on binary classification data sets.
		\change{
			In our findings we compare the chosen graph-based methods using all distance measures and observe that the results vary strongly with respect to the accuracy. As predicted by the ``no free lunch'' theorem, no clear best combination to employ in all cases is found. Our study provides a reproducible framework for future work in the direction of semi-supervised learning for time series with a focus on graph representations.}
	\end{abstract}
	\section{Introduction}
	Data are available more abundant than ever before and many of the observed processes produce data that are time-dependent. As a result the study of time series data is a subject of great importance \cite{fu2011review,bello2016social,chen2015data} and many different tasks are of interest depending on the particular application. The case of time series is interesting for tasks such as anomaly detection \cite{laptev2015generic}, motif computation \cite{chiu2003probabilistic} or time series forecasting \cite{de200625}. We refer to \cite{wei2006time,chatfield2019analysis,fawaz2019deep,abanda2019review} for more general introductions. 
	
	We here focus on the task of classification of time series \cite{wei2006semi,liao2005clustering,aghabozorgi2015time} in the context of semi-supervised learning \cite{zhu2009introduction,chapelle2009semi} where we want to label all data points based on the fact that only a small portion of the data is already pre-labeled.
	
	\begin{figure}[!htb]
		\begin{center}
			\includegraphics[width=.7\textwidth]{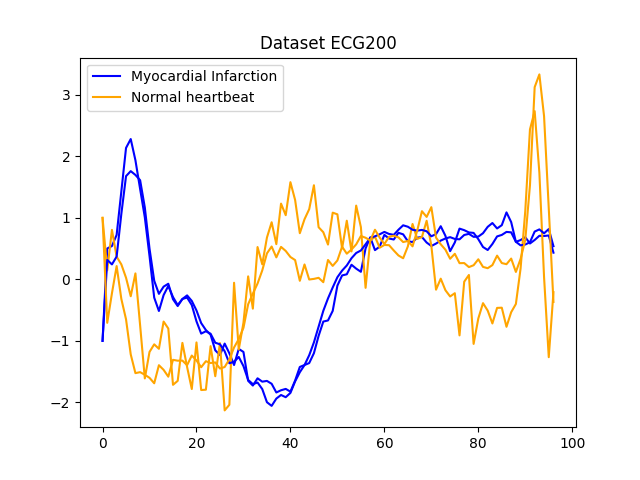}
			\caption{A typical example for time series classification. Given the dataset ECG200, the goal is to automatically separate all time series into the classes \textit{normal heartbeats} and \textit{myocardial infarction}. \label{fig::timeexample}}
		\end{center}
	\end{figure}
	
	An example is given in \cref{fig::timeexample} where we see some time series reflecting ECG data and the classification into normal heartbeats on the one hand and myocardial infarction on the other hand. In our applications, we assume that only for some of the time series the corresponding class is known a priori.
	Our focus is on the case when the data are incorporated into a graph. Each time series becomes a node within a weighted undirected graph and the edge-weight is proportional to the similarity between different time series. Graph-based approaches have become a standard tool in many learning tasks (cf. \cite{StollGAMM,Mercado:2019:ecmlpkdd,kipf2016semi,bertozzi2018uncertainty,spectral,bruna2013spectral} and the references mentioned therein). The matrix representation of the graph via its Laplacian \cite{chung1997spectral} leads to studying the network using matrix properties. An important ingredient in the construction of the Laplacian is the choice of the appropriate weight function. In many applications, the computation of the distance between time series or sub-sequences becomes a crucial task and this will be reflected in our choice of weight function. We consider several distance measures such as DTW \cite{dtw}, soft DTW \cite{cuturi2017soft}, and matrix profile \cite{gharghabi2020ultra}.
	
	We will embed these measures via the graph Laplacian into two different recently proposed semi-supervised learning frameworks. Namely, a diffuse interface approach that originates from material science \cite{bertozzi2012diffuse} via the graph Allen-Cahn equation as well as a method based on graph convolutional networks \cite{kipf2016semi}. Since these methods have originally been introduced outside of the field of time series learning, their relationship with time series distance measures has never been studied.
	Our goal is furthermore to compare these approaches with the well-known 1NN approach \cite{wei2006semi} and a simple optimization formulation solved relying on a linear system of equations. 
	\change{Our motivation follows that of \cite{bagnall2017great}, where many methods for supervised learning in the context of time series were compared, namely that we aim to provide a wide-ranging overview of recent methods based on a graph representation of the data and combined with several distance measures.}
	
	We structure the paper as follows. In Section \ref{sec::basics} we introduce some basic notations and make the case for transforming the data to graph form based on a motivation from unsupervised learning. In Section \ref{sec::dist} we discuss several distance measures with a focus on the well-known DTW measure as well as two recently emerged alternatives, i.e. Soft DTW and the MP distance. We use Section \ref{sec::ssl} to introduce the two semi-supervised learning methods in more detail, followed by a shorter description of their well-known competitors. Section \ref{sec::results} will allow us to compare the methods and study the hyperparameter selection.

	\section{Basics}
	\label{sec::basics}
\begin{figure}[bh]
		\subfloat[k-means \label{fig:kmeans}]{\includegraphics[width=.45\textwidth]{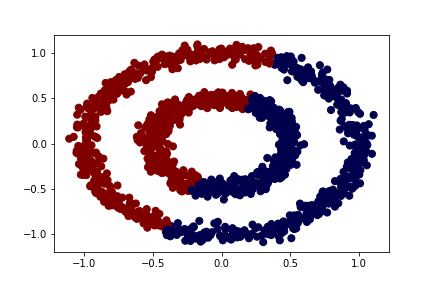}}
		\subfloat[Spectral Clustering\label{fig:spectral}]{\includegraphics[width=.45\textwidth]{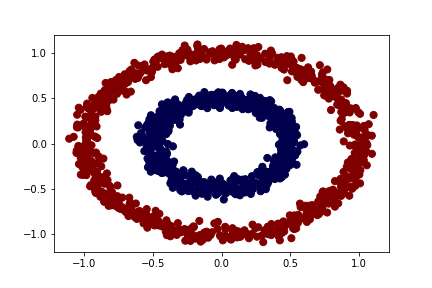}}
		\caption{Clustering of based on original data via k-means vs. transformed data via spectral clustering\label{fig:clustering}}
	\end{figure}

	We consider discrete time series $\bx_i$ given as a vector of real numbers of length $m_i$. In general, we allow for the time series to be of different dimensionality; later we often consider all $m_i=m$. We assume that we are given $n$ time series $\bx_i\in\R^{m_i}$. The goal of a classification task is to group the $n$ time series into a number $k$ of different \textit{clusters} $C_j$ with $j=1,\ldots,k$. In this paper we focus on the task of semi-supervised learning \cite{zhu2009introduction} where only some of the data are already labeled but we want to classify all available data simultaneously. Nevertheless, we review some techniques for unsupervised learning first as they deliver useful terminology.
	As such the \textit{k-means} algorithm is a prototype-based clustering algorithm that divides the given data into a predefined number of $k$ clusters \cite{macqueen1967some}. The idea behind $k$-means is rather simple as the cluster centroids are repeatedly updated and the data points are assigned to the nearest centroid until the centroids and data points have converged.  Often the termination condition is not handled that strictly. For example, the method can be terminated when only 1\% of the points change clusters. 
	The starting classes are often chosen at random but can also be assigned in a more systematic way by calculating the centers first and then assign the points to the nearest center. 
	While $k$-means remains very popular it also has certain weaknesses coming from its minimization of the sum of squared errors loss function \cite{mackay2003information}. We discuss this method in some detail here to point out the main mechanism and this is the measuring of all distances in the Euclidean norm, which would also be done when $k$-means is applied to time series. A simple two-dimensional example is shown in \cref{fig:clustering} where it is clear to see that $k$-means fails to capture the data manifold. In comparison, the alternative method shown, i.e., a spectral clustering technique, performs much better. We briefly discuss this method next as it forms the basis of the main techniques introduced in this paper.

	\subsection{Graph Laplacian and spectral clustering}
	As we illustrated in \cref{fig:clustering} the separation of the data into two-classes is rather difficult for $k$-means as the centroids are based on a $2$-norm minimization. One alternative to $k$-means is based on interpreting the data points as nodes in a graph. For this, we assume that we are given data points $x_{1},...,x_{n}$ and some measure of similarity \cite{spectral}. We define the weighted undirected {similarity graph} $G = (V,E)$ with the \textit{vertex} or \textit{node} set $V$ and the edge set $E$.
	We view the data points $\bx_i$ as vertices, $V=\{\bx_1, \ldots, \bx_n\}$, and if two nodes $(\bx_i, \bx_j)$ have a positive similarity function value, they are connected by an edge with weight $w_{ij}$ equal to that similarity.
	With this reformulation of the data we turn the clustering problem into a graph partitioning problem where we want to cut the graph into two or possibly more classes. This is usually done in such a way that the weight of the edges across the partition is minimal. 
	
	We collect all edge weights in the \textit{adjacency matrix} $W = (w_{ij})_{i,j=1,...,n}$. The degree of a vertex $\bx_i$ is defined as $d_i = \sum_{j=1}^{n} w_{ij}$ and the degree matrix $D$ is the diagonal matrix holding all $n$ node degrees. In our case we use a fully connected graph with the \textit{Gaussian similarity function}
	\begin{equation}
	w(\bx_{i},\bx_{j}) = \exp\Big(-\frac{\dist(\bx_i,\bx_j)^2}{\sigma^{2}}\Big),
	\label{eq::weight}
	\end{equation}
	where $\sigma$ is a scaling parameter and $\dist(\bx_i,\bx_j)$ is a particular distance function, such as the Euclidean distance  $\dist(\bx_i,\bx_j):=\Vert \bx_{i} - \bx_{j} \Vert^{2}$. Note that for similar nodes, the \textit{distance} function is small while the \textit{similarity} function is relatively large.
	
	We now use both the degree and weight matrix to define the \textit{graph Laplacian} as $L = D - W$.
	Often the \textit{symmetrically normalized Laplacian} defined via 
	\begin{equation}
	\Ls = D^{-\frac{1}{2}}LD^{-\frac{1}{2}} = I - D^{-\frac{1}{2}}WD^{-\frac{1}{2}}
	\label{eq:lsym}
	\end{equation}
	provides better clustering information \cite{spectral}. It has some very useful properties that we will exploit here. For example, given a non-zero vector $u \in \mathbb{R}^{n}$ we obtain the energy term
	\begin{equation}
	\begin{aligned}
	u^{\mathsf T}\Ls u = \frac{1}{2}\sum_{i,j} w_{ij} \left( \frac{u_{i}}{\sqrt{d_{i}}} - \frac{u_{j}}{\sqrt{d_{j}}}\right)^{2}.
	\end{aligned}
	\end{equation}
	Using this it is easy to see that $\Ls$ is positive semi-definite with non-negative eigenvalues $0 = \lambda_{1} \le \lambda_{2} \le ... \le \lambda_{n}$. The main advantage of the graph Laplacian is that based on its spectral information one can usually rely on transforming the data into a space where they are easier to separate \cite{chung1997spectral,spectral,belkin2001laplacian}. As a result one typically requires the spectral information corresponding to the smallest eigenvalues of $\Ls.$ The most famed eigenvector is the \textit{Fiedler vector}, i.e., the eigenvector corresponding to the first non-zero eigenvalue, which is bound to have a sign change and as a result can be used for binary classification. 
	The weight function \eqref{eq::weight} is also found in kernel methods \cite{shawe2004kernel,hofmann2008kernel} when the radial basis kernel is applied. 
	
	\subsection{Self-tuning}
	In order to improve the performance of the methods based on the graph Laplacian, tuning the parameter $\sigma$ is crucial. While hyperparameter tuning based on a grid search or cross validation is certainly possible we also consider a $\sigma$ that adapts to the given data. For spectral clustering, such a procedure was introduced in \cite{lihi}. Here we use this technique to learning with time series data. For each time series $\bx_i$ we assume a local scaling parameter $\sigma_i$. As a result, we have the generalized square distance as 
	\begin{equation}
	\begin{aligned}
	\frac{\dist(\bx_i,\bx_j)}{\sigma_i}\frac{\dist(\bx_i,\bx_j)}{\sigma_j}=\frac{\dist(\bx_i,\bx_j)^2}{\sigma_i\sigma_j}
	\end{aligned}
	\end{equation}
	and this gives the following adjacency matrix
	\begin{equation}
	\begin{aligned}
	W(i,j)= \exp\left(-\frac{\dist(\bx_i,\bx_j)^2}{\sigma_i\sigma_j}\right).
	\end{aligned}
	\end{equation}
	The authors in \cite{lihi} choose $\sigma_i$ as the distance to the $K$-th nearest neighbor of $\bx_i$ where $K$ is a fixed parameter, e.g., $K=9$ is used in \cite{bertozzi2012diffuse}.
	
	In Section \ref{sec::results} we will explore several different values for $K$ and their influence on the classification behavior.

	\section{Distance measures}
	\label{sec::dist}
	\change{
		We have seen so far that the Laplacian as well as typical kernel methods will rely on the choice of the distance measure $\dist(\bx_i,\bx_j)$. If all time series are of the same length then the easiest distance measure would be a Euclidean distance, which especially for large $n$ is fast to compute. This makes the Euclidean distance incredibly popular but it suffers from being sensitive to small shifts in the time series. As a result we discuss several popular and efficient methods for different distance measures. Our focus is to illustrate in an empirical study how the choice of distance measure impacts the performance of graph-based learning and it is clear that very likely there will be no clear winner in this competition but rather further insights for future research (cf. \cite{keogh2003need}).}
	
	\subsection{Dynamic Time Warping}
	
	We first discuss the distance measure of Dynamic Time Warping (DTW, \cite{dtw}). 
	By construction, DTW is an algorithm to find an optimal alignment between time series. 
	
	In the following, we adapt the notation of \cite{dtw} to our case. Consider two time series $\bx$ and $\bxt$ of lengths $m$ and $\tilde{m}$, respectively, with entries $x_i, \tilde{x}_i \in \mathbb{R}$ for $i=1,\ldots,m$ and $j=1,\ldots,\tilde{m}$. We obtain the local cost matrix $C \in \mathbb{R}^{m \times \tilde{m}}$ by assembling the local differences for each pair of elements, i.e., $C_{ij} = |x_i - \tilde{x}_j|$.
	
\begin{figure}[htbp]
		\centering
		\subfloat{\includegraphics[width=.45\textwidth]{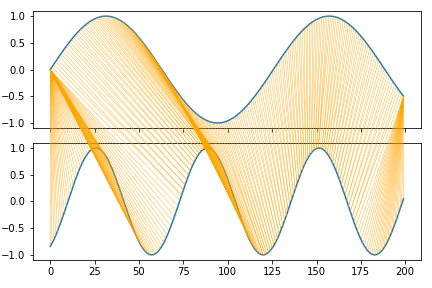}}
		\subfloat{\includegraphics[width=.45\textwidth]{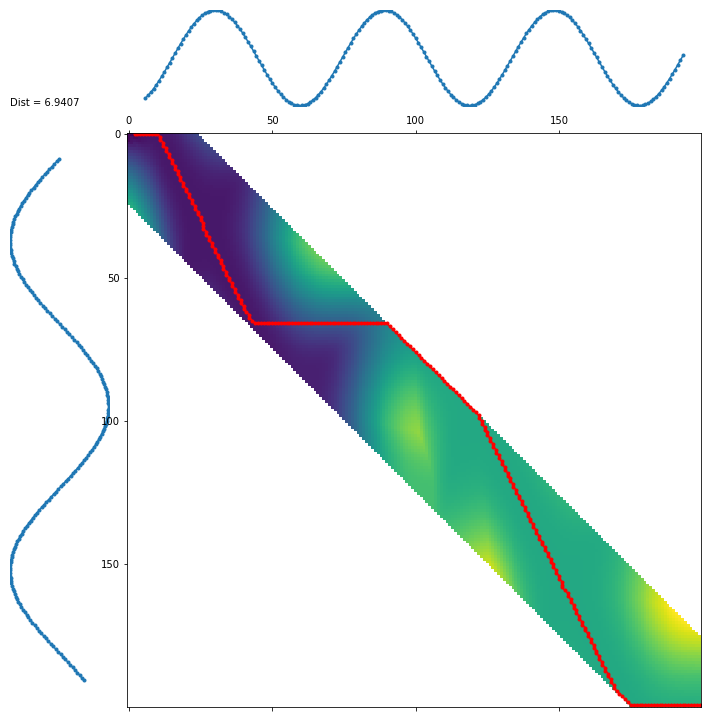}}
		\caption{DTW warping (left) and warpings paths (right) \label{fig:dtw}}
	\end{figure}

	The DTW distance is defined via \textit{$(m,\tilde{m})$-warping paths}, which are sequences of index tuples $p = \big((i_1,j_1), . . . , (i_L,j_L)\big)$ with boundary, monotonicity, and step size conditions
	\begin{gather*}
	1 = i_1 \leq i_2 \leq \ldots \leq i_L = m, \quad 1 = j_1 \leq j_2 \leq \ldots \leq \tilde{m}, \\
	(i_{l+1}-i_l, \; j_{l+1}-j_l) \in \lbrace (1, 0), (0, 1), (1, 1)\rbrace \quad (l = 1,\ldots, L - 1).
	\end{gather*}
	The total cost of such a path with respect to $\bx,\bxt$ is defined as
	\[
	c_p(\bx,\by) = \sum_{l=1}^{L} |x_{i_l} - \tilde{x}_{j_l}|.
	\]
	The DTW distance is then defined as the minimum cost of any warping path:
	\begin{equation}
	\DTW(\bx,\by):=\min \lbrace c_p(\bx,\by) \mid p \text{ is a $(m,\tilde{m})$-warping path} \rbrace.
	\end{equation}
	Both the warping and the warping path are illustrated in \cref{fig:dtw}.
	
	Computing the optimal warping path directly quickly becomes infeasible. However, we can use dynamic programming to evaluate the accumulated cost matrix $D$ recursively via
	\begin{equation}
	D(i,j) := |x_i - \tilde{x}_j| + \min \{ D(i,j-1), D(i-1,j), D(i-1,j-1) \}.
	\end{equation}
	The actual DTW distance is finally obtained as
	\begin{equation}
	\DTW(\bx,\by) = D(m,\tilde{m}).
	\end{equation}
	The DTW method is a heavily used distance measure for capturing the sometimes subtle similarities between time series. In the literature it is often stated that the computational cost of DTW are often described as too high. As a result one is interested in accelerating the DTW algorithm itself. One possibility arises from imposing additional {constraints} (cf. \cite{dtw,fastdtw}) such as the {Sakoe-Chiba Band} and the {Itakura parallelogram}. \change{While these are appealing concepts the authors in \cite{wu2020fastdtw} observe that the well-known FastDTW algorithm is in fact slower than DTW. As such, we rely on the implementation of DTW provided via \url{https://github.com/wannesm/dtaidistance}. We observe that for this implementation of DTW indeed FastDTW is outperformed frequently.}

	\subsection{Soft Dynamic Time Warping}
	
	Based on a slight reformulation of the above DTW scheme, we want to look at another time series distance measure, the \textit{Soft Dynamic Time Warping} (Soft DTW). It is an extension of DTW designed to allow differentiable introduced in ~\cite{blondel2020differentiable,cuturi2017soft}. We again start from the cost matrix $C$ with $C(i, j) = |x_i - \tilde{x}_j|$ for time series $\bx$ and $\bxt$. Each warping path can equivalently be described by a matrix $A \in \{0,1\}^{m\times\tilde{m}}$ with the following condition: The ones in $A$ form a path starting in $(1,1)$ going to $(m,\tilde{m})$, only using steps downwards, to the right and diagonal downwards. $A$ is called monotonic alignment matrix and we denote the set containing all these alignment matrices with $\mathcal{A}(m, \tilde{m})$. The Frobenius inner product $\langle A, C \rangle$ is then the sum of costs along the alignment $A$. Solving the following minimization problem leads us to a reformulation of the dynamic time warping introduced above as
	\begin{equation}   
	\text{DTW}(C) = \min_{A \in \mathcal{A}(N,M)} \langle A, C \rangle.
	\label{eq:dtwreformulation}  
	\end{equation} 
	With Soft DTW we involve all alignments possible in $\mathcal{A}(N, M)$ by replacing the minimization with a \textit{soft minimum}: 
	\begin{equation}   
	\min_{x \in S} f(x) \approx \mingamma_{x \in S} f(x) := - \gamma \log \sum_{x \in S} \exp\left(\frac{-f(x)}{\gamma}\right).
	\label{eq:softminimum}
	\end{equation}
	This function approximates the minimum of $f(x)$ and is differentiable. The parameter $\gamma$ controls the tuning between smoothness and approximation of the minimum. Using the DTW-function (\ref{eq:dtwreformulation}) within (\ref{eq:softminimum}) yields the expression for Soft Dynamic Time Warping written as 
	\begin{align*}
	\SDTWgamma(\bx,\bxt) &= \mingamma_{A \in \mathcal{A}(m, n)} \langle A, C \rangle \\ 
	&= - \gamma \log \sum_{A \in \mathcal{A}(m, n)} \exp\left(\frac{-\langle A, C \rangle}{\gamma}\right). \numberthis \label{eq:softdtw}
	\end{align*}
	This is now a differentiable alternative to DTW, which involves all alignments in our cost matrix. 
	
	Due to entropic bias, Soft DTW can generate negative values, which would cause issues for our use in time series classification.
	We apply the following remedy to overcome this drawback:
	\begin{equation}   
	\mathrm{Div}(\bx,\by)  =  \SDTWgamma(\bx,\by) -  \frac{1}{2}\cdot \big(\SDTWgamma(\bx,\bx) + \SDTWgamma(\by,\by)\big).
	\label{eq:softdtwdivergence}  
	\end{equation} 
	This measure is called {Soft DTW divergence} \cite{blondel2020differentiable} and will be employed in our experiments.

	\subsection{Matrix Profile Distance}
	
	Another alternative time series measure that has recently been introduced is the \textit{Matrix Profile Distance} (MP distance, \cite{gharghabi2020ultra}). This measure is designed for fast computation and finding similarities between time series.
	
	We will again introduce the concept of the matrix profile of two time series $\bx$ and $\bxt$. The matrix profile is based on the subsequences of these two time series. For a fixed window length $L$, the subsequence $\bx_{i,L}$ of a time series $\bx$ is defined as a contiguous $L$-element subset of $\bx$ via $\bx_{i,L} = (x_{i}, x_{i+1}, \ldots, x_{i+L-1})$. 
	The \textit{all-subsequences set} $A$ of $\bx$ contains all possible subsequences of $\bx$ with length $L$, $A = \{\bx_{1,L},  \bx_{2,L}, \ldots, \bx_{m - L + 1,L}\}$, where $m$ is again the length of $\bx$.
	
	For the matrix profile, we need the all-subsequences sets $A$ and $B$ of both time series $\bx$ and $\bxt$.
	The matrix profile $\mathbf{P}_{\mathrm{ABBA}}$ is the set consisting of the closest Euclidean distances from each subsequence in $A$ to any subsequence in $B$ and vice versa:
	\[
	\begin{aligned}
	\mathbf{P}_{\mathrm{ABBA}} &= 
	\left\{ \min_{\bxt_{j,L} \in B} \| \bx_{i,L} - \bxt_{j,L} \| \; \middle| \; \bx_{i,L} \in A \right\}
	\;\cup \\
	& \qquad \qquad
	\left\{ \min_{\bx_{i,L} \in A} \| \bxt_{j,L} - \bx_{i,L} \| \; \middle| \; \bxt_{j,L} \in B \right\}
	\end{aligned}
	\]
	With the matrix profile, we can finally define the MP distance based on the idea that two time series are similar if they have many similar subsequences. 
	We do not consider the smallest or the largest value of $\textbf{P}_{ABBA}$ because then the MP distance could be too rough or too detailed. For example, if we would have two rather similar time series, but either one has a noisy spike or some missing values, then the largest value of the matrix profile could give a wrong impression about the similarity of these two time series. Instead, the distance 
	is defined as
	\[ \MPdist(X,Y) = \text{$k$-th smallest value in sorted }  \textbf{P}_{ABBA}, \]
	where the parameter $k$ is typically set to $5\%$ of $2N$ \cite{gharghabi2020ultra}.
	
\begin{figure}[h]
		\begin{subfigure}{0.33\textwidth}
			\centering
			\includegraphics[width=\linewidth]{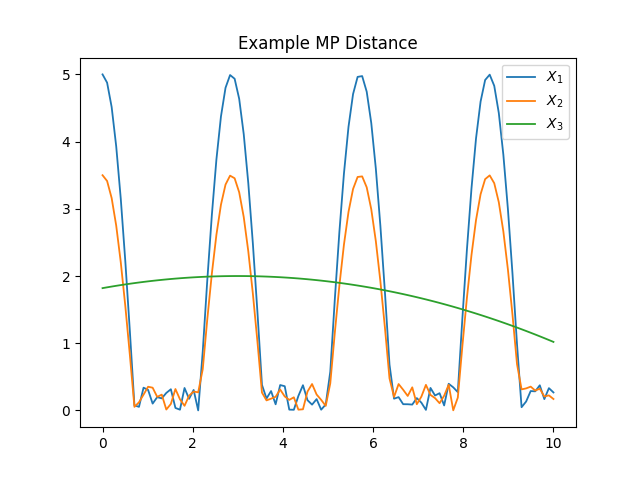}
			\caption{Time series $X_{1}$, $X_{2}$ and $X_{3}$}
			\label{fig:mpdist-example:time-series}
		\end{subfigure}%
		\begin{subfigure}{0.33\textwidth}
			\centering
			\includegraphics[width=\linewidth]{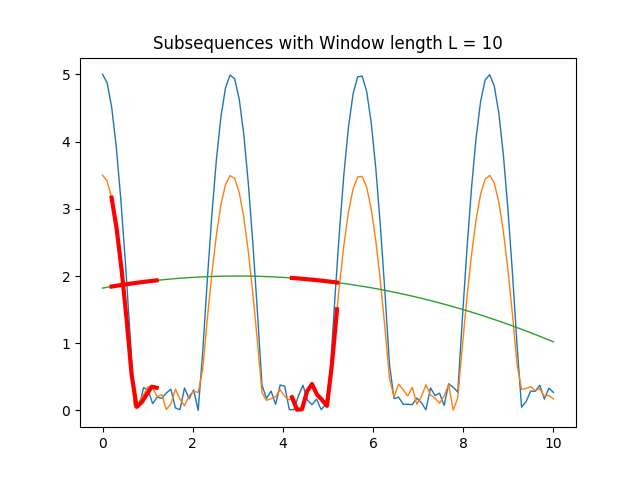}
			\caption{Smaller window length}
			\label{fig:mpdist-example:short-window-length}
		\end{subfigure}%
		\begin{subfigure}{0.33\textwidth}
			\includegraphics[width=\linewidth]{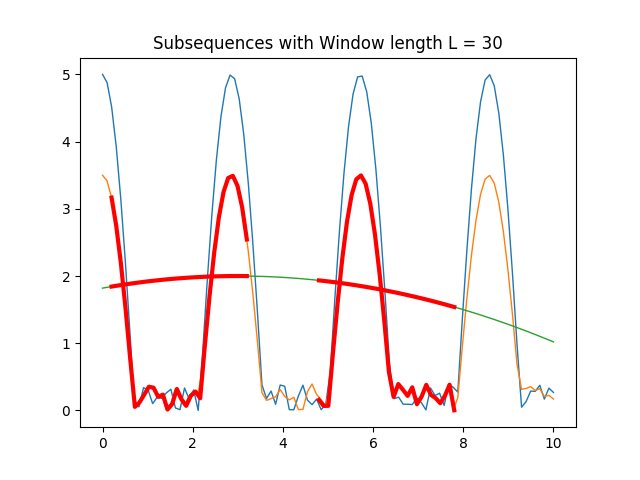}
			\caption{Larger window length}
			\label{fig:mpdist-example-long-window-length}
		\end{subfigure}
		\caption{Illustration of Matrix Profile distance, subsequences indicated in red}
		\label{fig:mpdist-example}
	\end{figure}

	We now illustrate the MP distance using an example as illustrated in \cref{fig:mpdist-example:time-series}, where we display three time series of length $N = 100.$ Our goal is to compare these time series using the MP distance. We observe that $X_{1}$ and $X_{2}$ have quite similar oscillations. The third time series $X_{3}$ does not share any obvious features with the first two sequences.
	
	The MP distance compares the subsequences of the time series, depending on the window length $L$. Choosing the window length to be $L = 40$, we get the following distances:
	\begin{equation*}
	\begin{split}
	\MPdist(X_{1}, X_{2}) & = 0.433, \\ \MPdist(X_{1}, X_{3}) & = 5.425, \\ \MPdist(X_{2}, X_{3}) & = 5.404.
	\end{split}
	\end{equation*}
	As we can see, the MP distance identified the similarity between $X_{1}$ and $X_{2}$ show the difference between the series $X_{1},X_{2}$ and $X_{3}$. We also want to show that the MP Distance depends on the window length $L$. Let us look at the MP distance between the lower oscillation time series $X_{2}$ and $X_{3}$, which is varying a lot for different values of $L$ as indicated in \cref{tab::mpdist}. Choosing $L = 10$ there is not a large portion of both time series to compare with and as a result we observe a small value for the MP distance, which does not describe the dissimilarity of $X_{2}$ and $X_{3}$ in a proper way. If we look at $L = 40$, there is a larger part of the time series structure to compare the two series. If there is a special recurring pattern in the time series, the length $L$ should be large enough to cover one recurrence. We illustrate the comparison based on different window lengths in \cref{fig:mpdist-example:short-window-length} and \ref{fig:mpdist-example-long-window-length}.
	\begin{table}
		\begin{center}
			\begin{tabular}{lcccc}
				L & 10 &  20 & 30 & 40 \\ \hline
				$\MPdist(X_{2}, X_{3})$ & 0.270 &  2,034 & 3,955 & 5,404 \\ 
			\end{tabular}
		\end{center}
		\caption{MP distance depending on the window length.\label{tab::mpdist}}
	\end{table}

	For the tests all data sets consist of time series with a certain length, varying for each data set. Thus we have to decide which window length $L$ should be chosen automatically in the classifier. An empirical study showed that choosing $L \approx N/2$ gives good classification results.
	
\begin{figure}
		\begin{center}
			\begin{tikzpicture}
			\begin{axis}[
			width=0.6\linewidth,
			height=0.4\linewidth,
			ymode=log,
			xmode=log,
			legend pos=outer north east,
			xmajorgrids, ymajorgrids,
			xmin=10,xmax=10000,
			ytick={1e-6,1e-4,1e-2,1,100},
			]
			
			\addplot[blue, thick] coordinates {
				(10,5.2753e-06) (20,6.7364e-06) (50,1.2287e-05) (100,4.4818e-05) (200,1.3264e-04) (500,4.9800e-04) (1000,1.1343e-03) (2000,4.3094e-03) (5000,2.5887e-02) (10000,6.1741e-02) (15000,2.4115e-01)
			};
			\addlegendentry{DTW (DTAI)}
			
			\addplot[darkgreen, thick] coordinates	
			{
				(10,3.3572e-04) (20,7.9262e-04) (50,2.2868e-03) (100,4.9275e-03) (200,1.0635e-02) (500,3.0314e-02) (1000,6.5708e-02) (2000,1.3918e-01) (5000,3.9371e-01) (10000,8.1406e-01) (15000,1.2455e+00)
			};
			\addlegendentry{DTW (\texttt{fastdtw})}
			
			\addplot[red, thick, dashed] coordinates {
				(10,1.3183e-04) (20,1.5021e-04) (50,2.4252e-04) (100,5.7394e-04) (200,2.1589e-03) (500,1.5725e-02) (1000,6.7018e-02) (2000,2.6396e-01) (5000,1.5331e+00) (10000,6.9020e+00) (15000,1.5945e+01)
			};
			\addlegendentry{SDTW}

			\addplot[amber, thick, dashdotted] coordinates 
			{
				(10,2.6579e-03) (20,2.6905e-03) (50,2.8309e-03) (100,2.9328e-03) (200,3.3305e-03) (500,4.4429e-03) (1000,7.2011e-03) (2000,1.7258e-02) (5000,6.7798e-02) (10000,8.6908e-01) (15000,1.9436e+00)	
			};
			\addlegendentry{MPDist}
			
			\addplot[black, thick, dotted] coordinates {
				(10,6.9693e-06) (20,7.0149e-06) (50,6.9124e-06) (100,6.9779e-06) (200,7.1861e-06) (500,7.2978e-06) (1000,7.7030e-06) (2000,8.1127e-06) (5000,1.4553e-05) (10000,2.2035e-05) (15000,2.8155e-05)
			};
			\addlegendentry{Euclidean}
			
			\end{axis}
			\end{tikzpicture}
		\end{center}
		\caption{Runtimes of distance computation between a single pair of time series with increasing length.
			\label{fig:runtime_distances}}
	\end{figure}
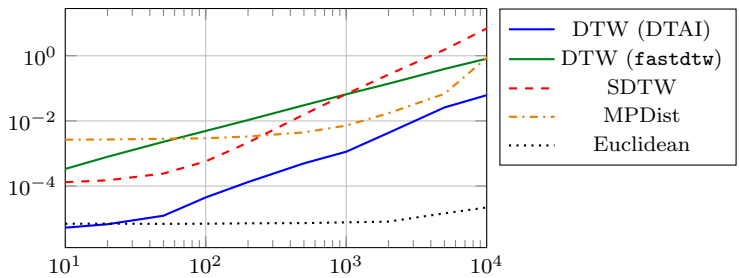

	\change{We briefly illustrate the computing times of the different distance measures when applied to time series of increasing length shown in \cref{fig:runtime_distances}. It can be seen that DTW is faster than fastDTW. Obviously, the Euclidean distance shows the best scalability. We also observe that the computation of the SDTW is scaling worse than the competing approaches when applied to longer time series.}
	
	\section{Semi-supervised learning based on graph Laplacians}
	\label{sec::ssl}
	In this section we propose the use of two methods that have recently gained wide attention. This first method is inspired by a partial differential equation model originating from material science and the second approach is based on neural networks that incorporate the graph structure of the labeled and unlabeled data.
	\subsection{Semi-supervised learning with phase field methods: Allen--Cahn model}
	Within the material science community phase field methods have been developed  to model the phase separation of a multicomponent alloy system (cf. \cite{TC94,AllC79}). The evolution of the phases over time is described by a partial differential equation (PDE) model, such as the Allen-Cahn \cite{AllC79} or Cahn-Hilliard equation \cite{cahn1958free} both non-linear reaction-diffusion equations of second and fourth order, respectively. These equations can be obtained as gradient flows of the Ginzburg--Landau energy functional
	$$
	\mathcal{E}(u)=\int \frac{\varepsilon}{2}\abs{\nabla{u}}^2+\frac{1}{\eps}\phi(u)
	$$
	where $u$ is the order parameter and $\eps$ a parameter reflecting the width of the interface between the pure phases. The polynomial $\phi$ is chosen to have minima at the pure phases, namely $u=-1$ and $u=1$, to enforce that a minimization of the Ginzburg--Landau energy will lead to phase separation. The Dirichlet energy term $\abs{\nabla{u}}^2$ corresponds to minimization of the interfacial length. The minimization is then performed using a gradient flow, which leads to the Allen-Cahn equation
	\begin{equation}
	u_t = \mathrm{\Delta} u -\frac{1}{\eps}\phi'(u)
	\label{eq::AC1}
	\end{equation}
	equipped with appropriate boundary and initial conditions. A modified Allen--Cahn equation was used for image inpainting, i.e. restoring damage parts in an image, where a misfit $\omega\left(f-u\right)$ term is added  to \cref{eq::AC1} (cf. \cite{BosKSW14,BerEG07}). Here, $\omega$ is a penalty parameter and $f$ is a function equal to the undamaged image parts or later training data. In \cite{bertozzi2012diffuse}, Bertozzi and Flenner extended this idea to the case of semi-supervised learning where the training data correspond to the undamaged image parts, i.e, the function $f$. Their idea is to consider the modified energy of the following form
	\begin{equation}
	\label{eq:graphenergy}
	E(u)=\frac{\eps}{2}u^T\Ls u+\frac{1}{4\eps}\sum_{i=1}^{n}(u_i^2-1)^2+\sum_{i=1}^{n}
	\frac{\omega_i}{2}(f_i-u_i)
	\end{equation}
	where $f_i$ holds the already assigned labels. Here, the first term in \eqref{eq:graphenergy} reflects the RatioCut based on the graph Laplacian, the second term enforces the pure phases, and the third term corresponds to incorporating the training data. Numerically, this system is solved using a convexity splitting approach \cite{bertozzi2012diffuse} where we write
	$$
	E(u)=E_1(u)-E_2(u)
	$$
	with 
	$$
	E_1(u):=\frac{\eps}{2}u^T\Ls u+\frac{c}{2}u^Tu
	$$
	and
	$$
	E_2(u):=\frac{c}{2}u^Tu-\frac{1}{4\eps}+\sum_{i=1}^{n}(u_i^2-1)^2-\sum_{i=1}^{n} \frac{\omega_i}{2}(f_i-u_i)
	$$
	where the positive parameter $c\in\R$ ensures convexity of both energies. In order to compute the minimizer of the above energy we use a gradient scheme where
	$$
	\frac{u^{l+1}-u^{l}}{\tau}=-\nabla E_1(u^{l+1})+\nabla E_2(u^{l})
	$$
	where the indices $k,\ k+1$ indicate the current and next time step, respectively. The variable $\tau$ is a hyperparameter but can be interpreted as a pseudo time-step. In more detail following the notation of \cite{Mercado:2019:ecmlpkdd}, this leads to
	$$
	\frac{u^{l+1}-u^{l}}{\tau}+\eps \Ls u^{l+1}+cu^{l+1}=cu^{l}-\frac{1}{\eps}\nabla \psi(u^{l})+\nabla \phi(u^{l}) 
	$$
	with
	$$
	\psi(u^l)=\sum_{i=1}^{n}((u^{l}_i)^2-1)^2,\quad \phi(u^{l})=\sum_{i=1}^{n} \frac{\omega_i}{2}(f_i-u^{l}_i).
	$$
	Expanding the order parameter in a number of the small eigenvectors $\phi_i$ of $\Ls$ via
	$u=\sum_{i=1}^{{m_e}}a_i\phi_i=\Phi_{m_e} a$
	where $a$ is a coefficient vector and $\Phi_{m_e}=[\phi_1,\ldots,\phi_{m_e}].$ This lets us arrive at
	$$
	(1+\eps\tau \lambda_j a^{l+1}_j+c\tau)a^{l+1}_j=(1+\tau c)a^{l}_j-\frac{1}{\eps}b_j^l+d^{l}_j,\quad \forall j=1,\ldots,{m_e}
	$$
	using
	$$
	b^l=\Phi_{m_e}^T\nabla \psi(\Phi_{m_e} a^l),\quad d^l=\Phi_{m_e}^T\nabla \phi(\Phi_{m_e} a^l).
	$$
	In \cite{garcia2014multiclass} the authors extend this to the case of multiple classes where again the spectral information of the graph Laplacian are crucial as the energy term includes
	$
	\frac{\eps}{2}\mathrm{trace}(U^T\Ls U)
	$
	with $U\in\R^{n,s}$, $s$ being the number of classes for segmentation. Details of the definition of the potential and the fidelity term incorporating the training data are found in \cite{garcia2014multiclass}.
	Further extensions of this approach have been suggested in \cite{Mercado:2019:ecmlpkdd,bosch2018generalizing,bergermann2020semi,budd2019graph,budd2020classification,calatroni2017graph,bertozzi2018uncertainty}.
	\subsection{Semi-supervised learning based on graph convolutional networks}
	Artificial neural networks and in particular deep neural networks have shown outstanding performance in many learning tasks \cite{goodfellow2016deep,lecun2015deep}. The incorporation of additional structural information via a graph structure has received wide attention \cite{bruna2013spectral} with particular success within the semi-supervised learning formulation \cite{kipf2016semi}. 
	
	Let $\mathbf{h}_i^{(l)}$ denote the hidden feature vector of the $i$-th node in the $l$-th layer. The feature mapping of a simple multilayer perceptron (MLP) computes the new features by multiplying with a weight matrix $\Theta^{(l)T}$ and adding a bias vector $b^{(l)}$, then applying a (potentially layer-dependent) ReLU activation function $\sigma_l$ in all layers except the last. This layer operation can be written as $\mathbf{h}_i^{l} = \sigma_l \Big( W^{(l)T} \mathbf{h}_i^{(l-1)} + b^{(l)} \Big)$.
	
	In Graph Neural Networks, the features are additionally propagated along the edges of the graph. This is achieved by forming weighted sums over the local neighborhood of each node, leading to
	\begin{equation}
	\mathbf{h}_i^{l} = \sigma_l \Big( \sum_{j \in \mathcal{N}_i \cup \{i\}} \frac{\hat{w}_{ij}}{\sqrt{\hat{d}_i \hat{d}_j}} \Theta^{(l)T} \mathbf{h}_j^{(l-1)} + b^{(l)} \Big).
	\end{equation}
	Here, $\mathcal{N}_i$ denotes the set of neighbors of node $i$, the $\hat{w}_{ij}$ denote the entries of the adjacency matrix $W$
	with added self loops, $\hat{W} = W + I$, and the $\hat{d}_i$ denote the row sums of that matrix. By adding the self loops, it is ensured that the original features of that node are maintained in the weighted sum.
	
	To obtain a matrix formulation, we can accumulate state matrices $X^{(l)}$ whose $n$ rows are the feature vectors $\mathbf{h}_i^{(l)T}$ for $i=1,\ldots,n$. The propagation scheme of a simple two-layer graph convolutional network can then be written as
	\begin{equation}
	\begin{aligned}
	X^{(1)} &= \sigma\Big( \hat{D}^{-1/2} \hat{W} \hat{D}^{-1/2} X^{(0)} \Theta^{(1)} + b^{(1)} \Big) \\
	X^{(2)} &= \hphantom{\sigma\Big(} \hat{D}^{-1/2} \hat{W} \hat{D}^{-1/2} X^{(1)} \Theta^{(2)} + b^{(2)}
	\end{aligned}
	\end{equation}
	where $\hat{D}$ is the diagonal matrix holding the $\hat{d}_i$.
	
	Multiplication with $\hat{D}^{-1/2} \hat{W} \hat{D}^{-1/2}$ can also be understood in a spectral sense as performing \emph{graph convolution} with the spectral filter function $\varphi(\lambda) = 1 - \lambda$. This means that the eigenvalues $\lambda$ of the graph Laplacian operator $\mathcal{L}$ (formed in this case \emph{after} adding the self loops) are transformed via $\varphi$ to obtain damping coefficients for the corresponding eigenvectors.
	
	It has been noted, e.g., in \cite{alfke21pinvgcn} that traditional graph neural networks including GCN are mostly targeted at the case of \emph{sparse} graphs, where each node is only connected to a small number of neighbors. The fully connected graphs that we utilize in this work present challenges for GCN through their spectral properties. Most notably, these \emph{dense} graphs typically have large eigengaps, i.e., the gap between the smallest eigenvalue $\lambda_1=0$ and the second eigenvalue $\lambda_2>0$ may be close to 1. Hence the GCN filter acts almost like a projection onto the undesirable eigenvector $\phi_1$. However, it has been observed in the same work that in some applications, GCNs applied to \emph{sparsified} graphs yield comparable results to dedicated dense methods. Our experiments justified only using Standard GCN on a $k$-nearest neighbor subgraph.
	

	\subsection{Other semi-supervised learning methods}
	In the context of graph-based semi-supervised learning a rather straightforward approach follows from minimizing the following objective
	\begin{equation}
	\min_{u}\frac{1}{2}\norm{u-f}_2^2+\frac{\beta}{2}u^T\Ls u
	\end{equation}
	where $f$ holds the values $1$, $-1$, and $0$ according to the labeled and unlabeled data. Calculating the derivative shows that in order to obtain $u$, we need to solve the following \textit{linear system} of equations
	$$
	\left(I+\beta \Ls\right)u=f
	$$
	where $I$ is the identity matrix of the appropriate dimensionality.
	
	Furthermore, we compare our previously introduced approaches to the well known one-nearest neighbor (1NN) method. In the context of time series classification this method was proposed in \cite{wei2006semi}.
	In each iteration, we identify the indices $i,j$ with the shortest distance between the labeled sample $\bx_i$ and the unlabeled sample $\bx_j$. The label of $\bx_i$ is then copied to $\bx_j$. This process is repeated until no unlabeled data remain.
	
	In \cite{xu2015time} the authors construct several graph Laplacians and then perform the semi-supervised learning based on a weighted sum of the Laplacian matrices.

	\section{Numerical experiments}
	\label{sec::results}
	In this section we illustrate how the algorithms discussed in this paper perform when applied to multiple time series data sets. \change{We here focus on binary classification and use time series taken from the UCR time series classification archive \footnotetext{We focussed on all binary classfication series listed in \texttt{TwoClassProblems.csv} within \url{http://www.timeseriesclassification.com/Downloads/Archives/Univariate2018_arff.zip}.} \cite{UCRArchive2018}.} All our codes are to be found at \url{https://github.com/dominikalfke/TimeSeriesSSL}.  The distance measure we use here are the previously introduced DTW, Soft DTW divergence, MP, and Euclidean distances.
	For completeness, we list the default parameters for all methods in \cref{tab:param}.
	

	\begin{table}[htb!]
		{\footnotesize
			\caption{Default parameters used in the experiments.\label{tab:param}}
			\begin{center}			
				\begin{tabular}{lp{9cm}}
					\toprule
					\textbf{Method} & \textbf{Parameters and default values} \\
					\midrule
					Allen--Cahn &  $m_e=20$, $\varepsilon=\frac{1}{\sqrt{n}}$, $c=\frac{3}{\varepsilon}+\omega,$ $\omega=1e10,$ $\tau=0.01,$ $tol=1e-8$ \\
					GCN &  10-NN sparsification, $h=32$, dropout $p=0.5$, \textsc{Adam} optimization \cite{adam}, learning rate 0.01, weight decay 0.0005, 500 epochs \\
					Linear System &  $\beta=1$, $tol=1e-5$ \\
					1NN & --- \\
					\bottomrule 
				\end{tabular}			
			\end{center}
		}
	\end{table}
	
	We split the presentation of the numerical results in the following way. We start by exploring the dependence of our schemes on some of the hyperparameters inherent in their derivation. We start by investigating the self-tuning parameters, namely the value of the chosen neighbor to compute the local scaling. We then study the performance of the Allen--Cahn model depending on the number of eigenpairs used for the approximation of the graph Laplacian. 
	For our main study, we pair up all distance measures with all learning methods and report the results on all datasets.
	Furthermore, we investigate how the method's performance depends on the number of available training data using random training splits.

	\subsection{Self-tuning values}
	In Section \ref{sec::basics} we proposed the use of the self-tuning approach for the Gaussian function within the weight matrix. The crucial hyperparameter we want to explore now is the choice of neighbor $k$ for the construction of $\sigma_i=\dist(\bx_i,\bx_{k,i})$ with $\bx_{k,i}$ the $k$-th nearest neighbor of the data point $\bx_i$. We can see from \cref{tab:ST} that the small values $k=7,20$ perform quite well in comparison to the larger self-tuning parameters. As a result we will use these smaller values in all further computations.

	\begin{table}
		\begin{tabular}{llccccc}
			\toprule
			\multicolumn{7}{c}{ECG200 $(n = 200)$}\\
			& & $k = 7$ & $k = 20$ & $k = \sqrt{n}$ & $k = 0.1n$ & $k = 0.05n$ \\
			\midrule
			\multirow{2}{*}{MPDist} & GCN & \textbf{83,58} \%  & 81,74 \%  & 81,90 \%  & 81,74 \%  & 82,54 \%  \\[0.3em]
			& Allen-Cahn& \textbf{81,00} \%  & 79,00 \%  & 80,00 \%  & 79,00 \%  & 80,00 \%  \\[0.3em]
			\multirow{2}{*}{SDTW} & GCN & \textbf{91,95} \%  & 91,34 \%  & 90,70 \%  & 91,43 \%  & 90,55 \%  \\
			& Allen-Cahn& \textbf{92,00} \%  & 90,00 \%  & 91,00 \%  & 90,00 \%  & 91,00  \%  \\[0.3em]
			\multirow{2}{*}{DTW} & GCN & 88,92 \%  & 86,76 \%  & 87,43 \%  & 86,76 \%  & \textbf{88,97} \%  \\
			& Allen-Cahn& 82,00 \%  & 82,00 \%  & \textbf{83,00} \%  & 82,00 \%  & 82,00  \%  \\
			\midrule
			\multicolumn{7}{c}{SonyAIBORobotSurface1 $(n = 621)$}\\
			& & $k = 7$ & $k = 20$ & $k = \sqrt{n}$ & $k = 0.1n$ & $k = 0.05n$ \\
			\midrule
			\multirow{2}{*}{MPDist} & GCN & \textbf{95,45} \%  & 88,74 \%  & 93,08 \%  & 78,10 \%  & 89,62 \%  \\
			& Allen-Cahn& \textbf{75,54} \%  & 72,88 \%  & 73,04 \%  & 75,37 \%  & 73,71 \%  \\[0.3em]
			\multirow{2}{*}{SDTW} & GCN & 90,32 \%  & 91,46 \%  & 92,48 \%  & 87,34 \%  & \textbf{92,85} \%  \\
			& Allen-Cahn& \textbf{93,68} \%  & 85,19 \%  & 82,36 \%  & 81,36 \%  & 82,36  \%  \\[0.3em]
			\multirow{2}{*}{DTW} & GCN & \textbf{97,59} \%  & 97,58 \%  & 97,48 \%  & 96,49 \%  & 97,35 \%  \\
			& Allen-Cahn& 84,03 \%  & 86,85 \%  & 87,69 \%  & 87,19 \%  & \textbf{88,19}  \%  \\
			\midrule
			\multicolumn{7}{c}{ECGFiveDays $(n = 884)$}\\
			& & $k = 7$ & $k = 20$ & $k = \sqrt{n}$ & $k = 0.1n$ & $k = 0.05n$ \\
			\midrule
			\multirow{2}{*}{MPDist} & GCN & 99,70 \%  & \textbf{99,77} \%  & 99,51 \%  & 99,66 \%  & 99,15 \%  \\
			& Allen-Cahn& 89,89 \%  & 90,71 \%  & 95,35 \%  & 95,82 \%  & \textbf{96,40} \%  \\[0.3em]
			\multirow{2}{*}{SDTW} & GCN & 97,30 \%  & 97,11 \%  & \textbf{97,31} \%  & 96,49 \%  & 97,06 \%  \\
			& Allen-Cahn& 82,00 \%  & 86,99 \%  & 85,48 \%  & 86,76 \%  & \textbf{87,57}  \%  \\[0.3em]
			\multirow{2}{*}{DTW} & GCN & 97,22 \%  & 97,19 \%  & \textbf{97,39} \%  & 97,20 \%  & 97,35 \%  \\
			& Allen-Cahn& \textbf{77,35} \%  & 76,31 \%  & 75,72 \%  & 73,17 \%  & 74,68 \%   \\
			\midrule
			\multicolumn{7}{c}{TwoLeadECG $(n = 1162)$}\\
			& & $k = 7$ & $k = 20$ & $k = \sqrt{n}$ & $k = 0.1n$ & $k = 0.05n$ \\
			\midrule
			\multirow{2}{*}{MPDist} & GCN & \textbf{99,81} \%  & 99,78 \%  & \textbf{99,81} \%  & 99,62 \%  & 99,74 \%  \\
			& Allen-Cahn& \textbf{99,12} \%  & 97,10 \%  & 96,49 \%  & 97,72 \%  & 96,57 \%  \\[0.3em]
			\multirow{2}{*}{SDTW} & GCN & \textbf{92,10} \%  & 90,74 \%  & 90,53 \%  & 89,98 \%  & 90,72 \%  \\
			& Allen-Cahn& \textbf{97,19} \%  & 93,24 \%  & 91,04 \%  & 87,27 \%  & 87,71 \%  \\[0.3em]
			\multirow{2}{*}{DTW} & GCN & 92,94 \%  & 94,04 \%  & 94,98 \%  & 93,97 \%  & \textbf{96,49} \%  \\
			& Allen-Cahn& 93,85 \%  & 92,36 \%  & 92,10 \%  & \textbf{94,12} \%  & 93,50 \%   \\
			\bottomrule
		\end{tabular}
		\caption{Study of self-tuning parameters.\label{tab:ST}}
	\end{table}

	\subsection{Spectral approximation}
	
	As described in Section \ref{sec::ssl} the Allen--Cahn equation is projected to a lower-dimensional space using the insightful information provided by the eigenvectors to the smallest eigenvalues of the graph Laplacian. We now investigate how the number of used eigenvectors impacts the accuracy.
	In the following we vary the number of eigenvalues from $10$ to $190$ and compare the performance of the Allen--Cahn method on three different datasets. The results are shown in \cref{tab:EVs} and it becomes clear that a vast number of eigenvectors does not lead to better classification accuracy. As a result we require a smaller number of eigenpair computations and also fewer computations within the Allen--Cahn scheme itself. The comparison was done for the self-tuning parameter $k=7$.

	\begin{table}
		\begin{center}
			\begin{tabular}{lccccc}
				\toprule
				\multicolumn{6}{c}{Dataset ECG200}\\
				Number of eigenvalues & 10 & 20 & 30 & 150 & 190 \\
				\midrule
				
				MPDist & 82.00 \% & 81.00 \% & \textbf{86.00} \% & 62.00 \% & 56.00 \% \\
				SDTW & 78.00 \% & \textbf{92.00} \% & \textbf{92.00} \% & 68.00 \% & 66.00 \% \\
				DTW & 78.00 \% & 82.00 \% & \textbf{87.00} \% & 69.00 \% & 54.00 \% \\
				\midrule
				
				\multicolumn{6}{c}{SonyAIBORobotSurface1}\\
				Number of eigenvalues & 10 & 20 & 30 & 500 & 600 \\
				\midrule
				
				MPDist & \textbf{85.36} \% & 75.54 \% & 73.04  \% & 51.58 \% & 51.08 \% \\
				SDTW & \textbf{96.17} \% & 93.68  \% & 83.19 \% & 52.08 \% & 49.92 \% \\
				DTW & \textbf{90.01} \% & 84.03 \% & 72.71 \% & 52.41 \% & 48.58 \% \\
				
				\midrule
				
				\multicolumn{6}{c}{ECGFiveDays}\\
				Number of eigenvalues & 10 & 20 & 30 & 700 & 800 \\
				\midrule
				MPDist & 87.19 \% & \textbf{89.89} \% & 85.95 \% & 50.29 \% & 51.22 \% \\
				SDTW & \textbf{91.52} \% & 82.00 \% & 84.20 \% & 54.00 \% & 52.38 \% \\
				DTW & 68.87 \% & \textbf{77.35} \% & 77.00 \% & 49.82 \% & 50.29 \% \\
				\bottomrule
			\end{tabular}
		\end{center}
		\caption{Varying the number of eigenpairs for the reduced Allen--Cahn equation.\label{tab:EVs}}
	\end{table}

	\subsection{Full method comparison}
\begin{figure}
		\begin{tikzpicture}
		\begin{axis}[
		width=\linewidth,
		height=2cm,
		at={(0,0.3cm)},
		anchor=south,
		hide axis,
		xmin=0, xmax=1, ymin=0, ymax=1,
		legend columns = -1,
		legend style={
			legend cell align=left, 
			at={(0.5,0.5)}, 
			anchor=center,
			/tikz/every even column/.append style={column sep=0.5cm}},
		]
		
		\addlegendimage{ybar, ybar legend, draw=plotColorNearestNeighbor, fill=plotColorNearestNeighbor}
		\addlegendentry{Nearest Neighbor}
		
		\addlegendimage{ybar, ybar legend, draw=plotColorLinearSystem, fill=plotColorLinearSystem}
		\addlegendentry{Linear System}
		
		\addlegendimage{ybar, ybar legend, draw=plotColorGCN, fill=plotColorGCN}
		\addlegendentry{GCN}
		
		\addlegendimage{ybar, ybar legend, draw=plotColorAllenCahn, fill=plotColorAllenCahn}
		\addlegendentry{Allen Cahn}
		
		\end{axis}
		
		\pgfplotsset{
			experiment bar chart/.style={
				ybar=0pt,
				width=0.32\textwidth,
				height=0.23\textwidth,
				anchor=north,
				ymin=50,
				ymax=100,
				xtick=data,
				xticklabels = {{SDTW}, MPDist, Eucl., DTW},
				ytick={50,60,70,80,90,100},
				yticklabel={\pgfmathprintnumber{\tick}\%},
				bar width={0.13cm},
				major x tick style = {draw=none},
				major y tick style = {draw=none},
				ymajorgrids,
				major grid style={draw=white, opacity=0.8}, 
				axis on top,
				xticklabel style={xshift=2pt, rotate=25, anchor=east},
				every tick label/.append style={font=\tiny}, 
				every x tick label/.append style={inner sep=0pt, outer sep=0pt},
				xlabel shift={-2pt},
			}
		}
		
		\foreach \x [count=\i from 1] in {-0.375,-0.125,0.125,0.375} {
			\foreach \y [count=\j from 1] in {0,1,...,5} {
				\pgfmathparse{0.2*\y}
				\coordinate (pos\i\j) at (\x\linewidth, -\pgfmathresult\linewidth);
			}
		}
		
		\input{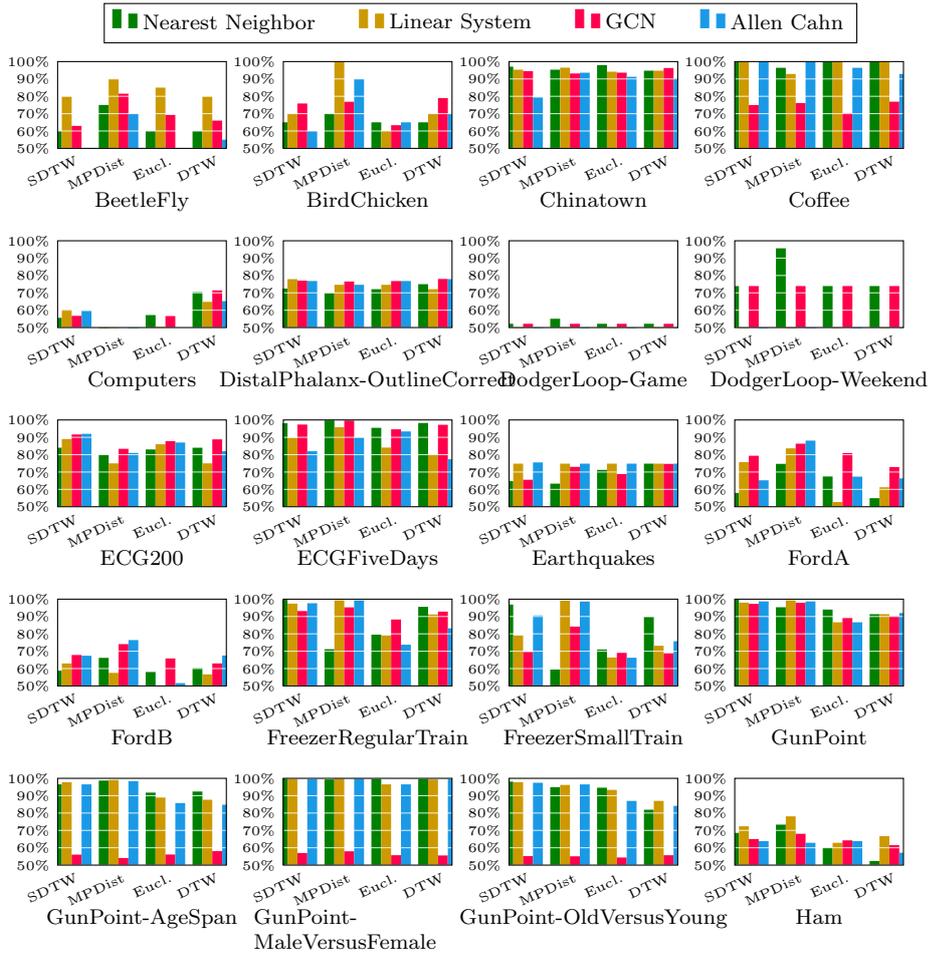}
		
		\end{tikzpicture}
		\caption{Comparison of the proposed methods using various distance measures for a variety of time series data.}
		\label{fig::comp1}
	\end{figure}

	{\color{black}
		We now compare the Allen-Cahn approach, the GCN scheme, the linear systems based method, and the 1NN algorithm, each paired up with each of the distance measures introduced in ~\cref{sec::dist}. Full results are listed in \cref{fig::comp1} and \cref{fig::comp2}. We show the comparison for all $42$ datasets.
		
		As can be seen there are several datasets where the performance of all methods is fairly similar even when the distance measure is varied. Here, we name \texttt{Chinatown}, \texttt{Earthquakes}, \texttt{GunPoint}, \texttt{ItalyPowerDemand}, \texttt{MoteStrain}, \texttt{Wafer}. There are several examples where the methods do not seem to perform well, with GCN and 1NN relatively similar outperforming the Linear System and Allen--Cahn approach. Such examples are \texttt{DodgerLoopGame}, \texttt{DodgerLoopWeekend}. The GCN method clearly does not perform well with the GunPoint datasets where the other methods clearly perform well. It is surprising to note that the Euclidean distance, given its computational speed and simplicity, does not come out as underperforming with respect to the accuracy across the different methods. There are very few datasets where one distance clearly outperforms the other choice. We name \texttt{ShapeletSim}, \texttt{ToeSegementation1} here.
	}

	\newcolumntype{R}{>{$}r<{$}} %
	\newcolumntype{V}[1]{>{[\;}*{#1}{R@{\;\;}}R<{\;]}} %

	\begin{figure}
		\begin{tikzpicture}
		\begin{axis}[
		width=\linewidth,
		height=2cm,
		at={(0,0.3cm)},
		anchor=south,
		hide axis,
		xmin=0, xmax=1, ymin=0, ymax=1,
		legend columns = -1,
		legend style={
			legend cell align=left, 
			at={(0.5,0.5)}, 
			anchor=center,
			/tikz/every even column/.append style={column sep=0.5cm}},
		]
		
		\addlegendimage{ybar, ybar legend, draw=plotColorNearestNeighbor, fill=plotColorNearestNeighbor}
		\addlegendentry{Nearest Neighbor}
		
		\addlegendimage{ybar, ybar legend, draw=plotColorLinearSystem, fill=plotColorLinearSystem}
		\addlegendentry{Linear System}
		
		\addlegendimage{ybar, ybar legend, draw=plotColorGCN, fill=plotColorGCN}
		\addlegendentry{GCN}
		
		\addlegendimage{ybar, ybar legend, draw=plotColorAllenCahn, fill=plotColorAllenCahn}
		\addlegendentry{Allen Cahn}
		
		\end{axis}
		
		\pgfplotsset{
			experiment bar chart/.style={
				ybar=0pt,
				width=0.32\textwidth,
				height=0.23\textwidth,
				anchor=north,
				ymin=50,
				ymax=100,
				xtick=data,
				xticklabels = {{SDTW}, MPDist, Eucl., DTW},
				ytick={50,60,70,80,90,100},
				yticklabel={\pgfmathprintnumber{\tick}\%},
				bar width={0.13cm},
				major x tick style = {draw=none},
				major y tick style = {draw=none},
				ymajorgrids,
				major grid style={draw=white, opacity=0.8}, 
				axis on top,
				xticklabel style={xshift=2pt, rotate=25, anchor=east},
				xlabel style={align=center, font=\small},
				every tick label/.append style={font=\tiny}, 
				every x tick label/.append style={inner sep=0pt, outer sep=0pt},
				xlabel shift={-2pt},
			}
		}
		
		\foreach \x [count=\i from 1] in {-0.375,-0.125,0.125,0.375} {
			\foreach \y [count=\j from 6] in {6,7,...,11} {
				\pgfmathparse{0.2*(\y-6)}
				\coordinate (pos\i\j) at (\x\linewidth, -\pgfmathresult\linewidth);
			}
		}
		
		\input{all_graphs2}
		
		\end{tikzpicture}
		\caption{Comparison of the proposed methods using various distance measures for a variety of time series data.}
		\label{fig::comp2}
	\end{figure}
	
	\subsection{Varying training splits}
	\change{
		In Fig.~\ref{fig::traincomp1}-\ref{fig::traincomp5} we vary the size of the training set from $1 \%$ to $20 \%$ of the available data. All reported numbers are averages over $100$ random splits. The numbers we observe mirror the performance of the full training size. We see that the methods show reduced performance when only $1$ \% of the training data are used but often reach an accuracy plateau when 5 to 10 \% of the training data are used. 
	}

	\newcommand{\splitgraphcaption}[1]{
		\caption{Method accuracy comparison for random training splits of different sizes (part #1/5)}
		\label{fig::traincomp#1}
	}
	\pgfplotsset{
		training size experiments/.style={
			width=0.3\textwidth,
			height=0.28\textwidth,
			anchor=center,
			ymin=50,
			ymax=102,
			xmin = 0,
			xmax = 21,
			xlabel style={font=\tiny},
			xtick=data,
			xtick={5,10,15,20},
			xticklabel={\pgfmathprintnumber{\tick}\%},
			ytick={50, 60, 70, 80, 90, 100},
			yticklabel={\pgfmathprintnumber{\tick}\%},
			major x tick style = {draw=none},
			major y tick style = {draw=none},
			xmajorgrids, ymajorgrids,
			every tick label/.append style={font=\tiny},
		},
		nearest neighbor/.style={
			plotColorNearestNeighbor, thick, mark=diamond, dashed, mark options={solid},
		},
		linear system/.style={
			plotColorLinearSystem, thick, mark=o, solid, mark options={solid},
		},
		standard gcn/.style={
			plotColorGCN, thick, mark=triangle, dashdotted, mark options={solid},
		},
		allen cahn/.style={
			plotColorAllenCahn, thick, mark=x, dotted, mark options={solid},
		},
	}
	\tikzset{
		distance label/.style={},
		dataset label/.style={rotate=90, anchor=center, font=\small, align=center},
	}
	\newcommand{\splitlegendaxis}{
		\begin{axis}[
			width=\linewidth,
			height=2cm,
			at={(0,0.12\linewidth)},
			anchor=south,
			hide axis,
			xmin=0, xmax=1, ymin=0, ymax=1,
			legend columns = -1,
			legend style={
				legend cell align=left, 
				at={(0.5,0.5)}, 
				anchor=center,
				/tikz/every even column/.append style={column sep=0.5cm}},
			]
			
			\addlegendimage{nearest neighbor}
			\addlegendentry{Nearest Neighbor}
			
			\addlegendimage{linear system}
			\addlegendentry{Linear System}
			
			\addlegendimage{standard gcn}
			\addlegendentry{GCN}
			
			\addlegendimage{allen cahn}
			\addlegendentry{Allen Cahn}
			
		\end{axis}
	}
	
	\input{all_split_graphs_new}

	\section{Conclusion}
	In this paper we took to the task of classifying time series data in a semi-supervised learning setting. For this we proposed to represent the data as a fully-connected graph where the edge weights are created based on a Gaussian similarity measure. The heart of this function is the difference measure between the time series, for which we used the (Soft) Dynamic Time Warping and Matrix Profile based distance measures as well as the Euclidean distance. We then investigated several learning algorithms, namely, the Allen--Cahn-based method, the Graph Convolutional Network scheme, and a linear system approach, all reliant on the graph Laplacian, as well as the Nearest Neighbor method. We then illustrated the performance of all pairs of distance measure and learning methods. \change{In this empirical study we observed that the methods tend to show an increased performance adding more training data. 	
		Studying all binary time-series with the \url{timeseriesclassification.com} repository gives results that in accordance with the no free lunch theorem show no clear winner. On the positive side the methods often perform quite well and there are only a few datasets with decreased performance. The comparison of the distance measures indicates there are certain cases where they outperform their competitors but also there is no clear winner with regards to accuracy. We believe that this empirical, reproducible study will encourage further research in this direction.}
	\section*{Acknowledgments}
	All authors would like to acknowledge the hard work and dedication by the team maintaining \url{www.timeseriesclassification.com/}.
	M. Stoll and L. Peroche both acknowledge the funding of the BMBF grant 01|S20053A. D. Alfke was partially supported by KINTUC project (S\"achsische Aufbaubank--Förderbank--(SAB) 100378180)

	
	\printbibliography 

@inproceedings{macqueen1967some,
	title={Some methods for classification and analysis of multivariate observations},
	author={MacQueen, James and others},
	booktitle={Proceedings of the fifth Berkeley symposium on mathematical statistics and probability},
	volume={1},
	number={14},
	pages={281--297},
	year={1967},
	organization={Oakland, CA, USA}
}

@article{wu2020fastdtw,
	title={FastDTW is approximate and Generally Slower than the Algorithm it Approximates},
	author={Wu, Renjie and Keogh, Eamonn J},
	journal={IEEE Transactions on Knowledge and Data Engineering},
	year={2020},
	publisher={IEEE}
}

@article{bagnall2017great,
	title={The great time series classification bake off: a review and experimental evaluation of recent algorithmic advances},
	author={Bagnall, Anthony and Lines, Jason and Bostrom, Aaron and Large, James and Keogh, Eamonn},
	journal={Data mining and knowledge discovery},
	volume={31},
	number={3},
	pages={606--660},
	year={2017},
	publisher={Springer}
}

@article{keogh2003need,
	title={On the need for time series data mining benchmarks: a survey and empirical demonstration},
	author={Keogh, Eamonn and Kasetty, Shruti},
	journal={Data Mining and knowledge discovery},
	volume={7},
	number={4},
	pages={349--371},
	year={2003},
	publisher={Springer}
}

@article {StollGAMM,
	AUTHOR = {Stoll, Martin},
	TITLE = {A literature survey of matrix methods for data science},
	JOURNAL = {GAMM-Mitt.},
	FJOURNAL = {GAMM-Mitteilungen},
	VOLUME = {43},
	YEAR = {2020},
	NUMBER = {3},
	PAGES = {e202000013, 4},
	ISSN = {0936-7195},
	MRCLASS = {65F99 (15A23 62R07 68T05 68W20)},
	MRNUMBER = {4166336},
}

@book{mackay2003information,
	title={Information theory, inference and learning algorithms},
	author={MacKay, David JC and Mac Kay, David JC},
	year={2003},
	publisher={Cambridge university press}
}

@article{fu2011review,
	title={A review on time series data mining},
	author={Fu, Tak-chung},
	journal={Engineering Applications of Artificial Intelligence},
	volume={24},
	number={1},
	pages={164--181},
	year={2011},
	publisher={Elsevier}
}

@incollection{wei2006time,
	title={Time series analysis},
	author={Wei, William WS},
	booktitle={The Oxford Handbook of Quantitative Methods in Psychology: Vol. 2},
	year={2006}
}

@article{abanda2019review,
	title={A review on distance based time series classification},
	author={Abanda, Amaia and Mori, Usue and Lozano, Jose A},
	journal={Data Mining and Knowledge Discovery},
	volume={33},
	number={2},
	pages={378--412},
	year={2019},
	publisher={Springer}
}

@article{bello2016social,
	title={Social big data: Recent achievements and new challenges},
	author={Bello-Orgaz, Gema and Jung, Jason J and Camacho, David},
	journal={Information Fusion},
	volume={28},
	pages={45--59},
	year={2016},
	publisher={Elsevier}
}

@book{chatfield2019analysis,
	title={The analysis of time series: an introduction with {R}},
	author={Chatfield, Chris and Xing, Haipeng},
	year={2019},
	publisher={CRC press}
}

@inproceedings{wei2006semi,
	title={Semi-supervised time series classification},
	author={Wei, Li and Keogh, Eamonn},
	booktitle={Proceedings of the 12th ACM SIGKDD international conference on Knowledge discovery and data mining},
	pages={748--753},
	year={2006}
}

@article{liao2005clustering,
	title={Clustering of time series data -- a survey},
	author={Liao, T Warren},
	journal={Pattern recognition},
	volume={38},
	number={11},
	pages={1857--1874},
	year={2005},
	publisher={Elsevier}
}

@article{aghabozorgi2015time,
	title={Time-series clustering -- a decade review},
	author={Aghabozorgi, Saeed and Shirkhorshidi, Ali Seyed and Wah, Teh Ying},
	journal={Information Systems},
	volume={53},
	pages={16--38},
	year={2015},
	publisher={Elsevier}
}

@article{gharghabi2020ultra,
	title={An ultra-fast time series distance measure to allow data mining in more complex real-world deployments},
	author={Gharghabi, Shaghayegh and Imani, Shima and Bagnall, Anthony and Darvishzadeh, Amirali and Keogh, Eamonn},
	journal={Data Mining and Knowledge Discovery},
	volume={34},
	pages={1104--1135},
	year={2020},
	publisher={Springer}
}

@book{zhu2009introduction,
	title={Introduction to semi-supervised learning},
	author={Zhu, Xiaojin and Goldberg, Andrew},
	year={2009},
	publisher={Morgan \& Claypool Publishers}
}

@article{chapelle2009semi,
	title={Semi-supervised learning},
	author={Chapelle, Olivier and Sch{\"o}lkopf, Bernhard and Zien, Alexander},
	journal={IEEE Transactions on Neural Networks},
	volume={20},
	number={3},
	pages={542--542},
	year={2009},
	publisher={IEEE}
}

@article{chen2015data,
	title={Data mining for the internet of things: literature review and challenges},
	author={Chen, Feng and Deng, Pan and Wan, Jiafu and Zhang, Daqiang and Vasilakos, Athanasios V and Rong, Xiaohui},
	journal={International Journal of Distributed Sensor Networks},
	volume={11},
	number={8},
	pages={431047},
	year={2015},
	publisher={SAGE Publications Sage UK: London, England}
}

@ARTICLE{TC94,
	author = {Taylor, Jean E. and Cahn, John W.},
	title = {Linking anisotropic sharp and diffuse surface motion laws via gradient flows},
	journal = {J. Statist. Phys.},
	year = {1994},
	volume = {77},
	pages = {183--197},
	number = {1-2},
	coden = {JSTPSB},
	fjournal = {Journal of Statistical Physics},
	issn = {0022-4715},
	mrclass = {58E12 (49Q05 53C42)},
	mrnumber = {MR1300532 (95j:58029)},
	mrreviewer = {Nathan Smale},
}

@article{AllC79,
	title={A microscopic theory for antiphase boundary motion and its application to antiphase domain coarsening},
	author={Allen, S.M. and Cahn, J.W.},
	journal={Acta Metallurgica},
	volume={27},
	number={6},
	pages={1085--1095},
	year={1979},
	publisher={Elsevier}
}

@article{cahn1958free,
	title={Free energy of a nonuniform system. I. Interfacial free energy},
	author={Cahn, John W and Hilliard, John E},
	journal={The Journal of chemical physics},
	volume={28},
	number={2},
	pages={258--267},
	year={1958},
	publisher={American Institute of Physics}
}

@article{BerEG07,
	title={Inpainting of binary images using the {C}ahn-{H}illiard equation},
	author={Bertozzi, A.L. and Esedoglu, S. and Gillette, A.},
	journal={Image Processing, IEEE Transactions on},
	volume={16},
	number={1},
	pages={285--291},
	year={2007},
	publisher={IEEE}
}

@article{garcia2014multiclass,
	title={Multiclass data segmentation using diffuse interface methods on graphs},
	author={Garcia-Cardona, Cristina and Merkurjev, Ekaterina and Bertozzi, Andrea L and Flenner, Arjuna and Percus, Allon G},
	journal={IEEE transactions on pattern analysis and machine intelligence},
	volume={36},
	number={8},
	pages={1600--1613},
	year={2014},
	publisher={IEEE}
}

@article{bergermann2020semi,
	title={Semi-supervised Learning for Multilayer Graphs Using Diffuse Interface Methods and Fast Matrix Vector Products},
	author={Bergermann, Kai and Stoll, Martin and Volkmer, Toni},
	journal={
	SIAM Journal on Mathematics of Data Science, accepted},
	year={2021}
}

@article{budd2019graph,
	title={Graph {MBO} as a semi-discrete implicit {E}uler scheme for graph {A}llen-{C}ahn},
	author={Budd, Jeremy and van Gennip, Yves},
	journal={arXiv preprint arXiv:1907.10774},
	year={2019}
}

@article{budd2020classification,
	title={Classification and image processing with a semi-discrete scheme for fidelity forced {A}llen-{C}ahn on graphs},
	author={Budd, Jeremy and van Gennip, Yves and Latz, Jonas},
	journal={arXiv preprint arXiv:2010.14556},
	year={2020}
}

@article{calatroni2017graph,
	title={Graph clustering, variational image segmentation methods and {H}ough transform scale detection for object measurement in images},
	author={Calatroni, Luca and van Gennip, Yves and Sch{\"o}nlieb, Carola-Bibiane and Rowland, Hannah M and Flenner, Arjuna},
	journal={Journal of Mathematical Imaging and Vision},
	volume={57},
	number={2},
	pages={269--291},
	year={2017},
	publisher={Springer}
}

@article{blondel2020differentiable,
	title={Differentiable Divergences Between Time Series},
	author={Blondel, Mathieu and Mensch, Arthur and Vert, Jean-Philippe},
	journal={arXiv preprint arXiv:2010.08354},
	year={2020}
}

@article{fawaz2019deep,
	title={Deep learning for time series classification: a review},
	author={Fawaz, Hassan Ismail and Forestier, Germain and Weber, Jonathan and Idoumghar, Lhassane and Muller, Pierre-Alain},
	journal={Data Mining and Knowledge Discovery},
	volume={33},
	number={4},
	pages={917--963},
	year={2019},
	publisher={Springer}
}

@book{goodfellow2016deep,
	title={Deep learning},
	author={Goodfellow, Ian and Bengio, Yoshua and Courville, Aaron and Bengio, Yoshua},
	volume={1},
	number={2},
	year={2016},
	publisher={MIT press Cambridge}
}

@article{lecun2015deep,
	title={Deep learning},
	author={LeCun, Yann and Bengio, Yoshua and Hinton, Geoffrey},
	journal={nature},
	volume={521},
	number={7553},
	pages={436--444},
	year={2015},
	publisher={Nature Publishing Group}
}

@article{bruna2013spectral,
	title={Spectral networks and locally connected networks on graphs},
	author={Bruna, Joan and Zaremba, Wojciech and Szlam, Arthur and LeCun, Yann},
	journal={arXiv preprint arXiv:1312.6203},
	year={2013}
}

@article{kipf2016semi,
	title={Semi-supervised classification with graph convolutional networks},
	author={Kipf, Thomas N and Welling, Max},
	journal={arXiv preprint arXiv:1609.02907},
	year={2016}
}

@article{bertozzi2018uncertainty,
	title={Uncertainty quantification in graph-based classification of high dimensional data},
	author={Bertozzi, Andrea L and Luo, Xiyang and Stuart, Andrew M and Zygalakis, Konstantinos C},
	journal={SIAM/ASA Journal on Uncertainty Quantification},
	volume={6},
	number={2},
	pages={568--595},
	year={2018},
	publisher={SIAM}
}

@article{bosch2018generalizing,
	title={Generalizing diffuse interface methods on graphs: nonsmooth potentials and hypergraphs},
	author={Bosch, Jessica and Klamt, Steffen and Stoll, Martin},
	journal={SIAM Journal on Applied Mathematics},
	volume={78},
	number={3},
	pages={1350--1377},
	year={2018},
	publisher={SIAM}
}

@InProceedings{Mercado:2019:ecmlpkdd,
	title = {Node Classification for Signed Social Networks Using Diffuse Interface Methods},
	author = {Mercado, Pedro and Bosch, Jessica and Stoll, Martin},
	booktitle = {ECMLPKDD},
	month = {September},
	year = {2019}
}

@Article{BosKSW14,
	author = {Jessica Bosch and David Kay and Martin Stoll and Andy Wathen},
	title = {Fast Solvers for {C}ahn-{H}illiard Inpainting},
	journal = {SIAM Journal on Imaging Sciences},
	year = {2014},
	OPTkey = {},
	volume = {7},
	issue = {1},
	OPTnumber = {},
	pages = {67--97},
	OPTmonth = {},
	OPTnote = {},
	OPTannote = {},
	OPTurl = {},
	OPTdoi = {},
	OPTissn = {},
	OPTlocalfile = {},
	OPTabstract = {}
}

@inproceedings{laptev2015generic,
	title={Generic and scalable framework for automated time-series anomaly detection},
	author={Laptev, Nikolay and Amizadeh, Saeed and Flint, Ian},
	booktitle={Proceedings of the 21th ACM SIGKDD international conference on knowledge discovery and data mining},
	pages={1939--1947},
	year={2015}
}

@inproceedings{chiu2003probabilistic,
	title={Probabilistic discovery of time series motifs},
	author={Chiu, Bill and Keogh, Eamonn and Lonardi, Stefano},
	booktitle={Proceedings of the ninth ACM SIGKDD international conference on Knowledge discovery and data mining},
	pages={493--498},
	year={2003}
}

@article{de200625,
	title={25 years of time series forecasting},
	author={De Gooijer, Jan G and Hyndman, Rob J},
	journal={International journal of forecasting},
	volume={22},
	number={3},
	pages={443--473},
	year={2006},
	publisher={Elsevier}
}

@book{chung1997spectral,
	title={Spectral graph theory},
	author={Chung, Fan RK and Graham, Fan Chung},
	number={92},
	year={1997},
	publisher={American Mathematical Soc.}
}

@inproceedings{xu2015time,
	title={Time series analysis with graph-based semi-supervised learning},
	author={Xu, Zhao and Funaya, Koichi},
	booktitle={2015 IEEE International Conference on Data Science and Advanced Analytics (DSAA)},
	pages={1--6},
	year={2015},
	organization={IEEE}
}

@book{shawe2004kernel,
	title={Kernel methods for pattern analysis},
	author={Shawe-Taylor, John and Cristianini, Nello and others},
	year={2004},
	publisher={Cambridge university press}
}

@article{bertozzi2012diffuse,
	title={Diffuse interface models on graphs for classification of high dimensional data},
	author={Bertozzi, Andrea L and Flenner, Arjuna},
	journal={Multiscale Modeling \& Simulation},
	volume={10},
	number={3},
	pages={1090--1118},
	year={2012},
	publisher={SIAM}
}

@article{hofmann2008kernel,
	title={Kernel methods in machine learning},
	author={Hofmann, Thomas and Sch{\"o}lkopf, Bernhard and Smola, Alexander J},
	journal={The annals of statistics},
	pages={1171--1220},
	year={2008},
	publisher={JSTOR}
}

@article{belkin2001laplacian,
	title={Laplacian eigenmaps and spectral techniques for embedding and clustering},
	author={Belkin, Mikhail and Niyogi, Partha},
	journal={Advances in neural information processing systems},
	volume={14},
	pages={585--591},
	year={2001}
}

@article{spectral,
author = {von Luxburg, Ulrike},
title = {A tutorial on Spectral Clustering},
journal = {Statistics and Computing 17 (4)},
pages = {395-416},
year={2007}
}

@inproceedings{cuturi2017soft,
	title={{Soft-DTW}: a differentiable loss function for time-series},
	author={Cuturi, Marco and Blondel, Mathieu},
	booktitle={International Conference on Machine Learning},
	pages={894--903},
	year={2017},
	organization={PMLR}
}

@book{dtw,
	title={Information retrieval for music and motion},
	author={M{\"u}ller, Meinard},
	volume={2},
	year={2007},
	publisher={Springer}
}

@article{UCRArchive2018,
  title={The UCR time series archive},
author={Dau, Hoang Anh and Bagnall, Anthony and Kamgar, Kaveh and Yeh, Chin-Chia Michael and Zhu, Yan and Gharghabi, Shaghayegh and Ratanamahatana, Chotirat Ann and Keogh, Eamonn},
journal={IEEE/CAA Journal of Automatica Sinica},
volume={6},
number={6},
pages={1293--1305},
year={2019},
publisher={IEEE}
}

@article{fastdtw,
author = {Salvador, Stan and Chan, Philip K.},
title = {Toward Accurate Dynamic Time Warping in Linear Time and Space},
journal = {Intelligent Data Analysis 11(5)},
pages = {70-80},
year={2004}
}

@inproceedings{lihi,
	title={Self-tuning spectral clustering},
	author={Zelnik-Manor, Lihi and Perona, Pietro},
	booktitle={Advances in neural information processing systems},
	pages={1601--1608},
	year={2005}
}

@article{alfke21pinvgcn,
	title={Pseudoinverse Graph Convolutional Networks: Fast Filters Tailored for Large Eigengaps of Dense Graphs and Hypergraphs},
	author={Alfke, Dominik and Stoll, Martin},
	journal={Data Mining and Knowledge Discovery, accepted},
	year={2021}
}

@inproceedings{adam,
	title = {Adam: A Method for Stochastic Optimization},
	_booktitle = {Proceedings of the 3rd International Conference on Learning Representations},
	booktitle = {Proc Int Conf Learn Represent},
	year = {2015},
	author = {Kingma, Diederik and Ba, Jimmy Lei},
	series = {ICLR ’15},
}
\end{document}